\documentclass[letterpaper]{article} 
\usepackage{ijcai20}  
\usepackage{times}  
\usepackage{helvet} 
\usepackage{courier}  
\usepackage[hyphens]{url}  
\usepackage{graphicx} 
\usepackage{graphicx}  
\usepackage{bbold,epstopdf}
\usepackage{url,graphicx,tabularx,array, datetime, booktabs, mathrsfs, color, tablefootnote,tabu} 
\usepackage{paralist}
\usepackage{helvet}
\usepackage{float}
\usepackage{amsmath, bm, amsfonts, cases, enumitem, booktabs}
\usepackage{algorithm}
\usepackage{algpseudocode}
\title{Memory Augmented Neural Model for Incremental Session-based Recommendation}

\author{Fei Mi \and Boi Faltings \\
\affiliations
Artificial Intelligence Laboratory, \'Ecole Polytechnique F\'ed\'erale de Lausanne (EPFL)\\
\emails
fei.mi@epfl.ch,
boi.faltings@epfl.ch
}

\begin{document}

      \maketitle
      
      \begin{abstract}
            
 Increasing concerns with privacy have stimulated interests in {\em 
Session-based Recommendation} (SR) using no personal data other than what is observed in the current browser session.
Existing methods are evaluated in static settings which rarely occur in real-world applications.
To better address the dynamic nature of \textit{SR} tasks, we study an \textit{incremental SR} scenario, where new items and preferences appear continuously. 
We show that existing neural recommenders can be used in \textit{incremental SR} scenarios with \textit{small} incremental updates to alleviate computation overhead and catastrophic forgetting.
More importantly, we propose a general framework called \textbf{M}emory \textbf{A}ugmented \textbf{N}eural model~(\texttt{MAN}). \texttt{MAN} augments a base neural recommender with a continuously queried and updated \textit{nonparametric} memory, and the predictions from the neural and the memory components are combined through another lightweight gating network.
We empirically show that \texttt{MAN} is well-suited for the incremental \textit{SR} task, and it consistently outperforms state-of-the-art neural and nonparametric methods. We analyze the results and demonstrate that it is particularly good at incrementally learning preferences on new and infrequent items.
            
      \end{abstract}

      \section{Introduction}

      Due to new privacy regulations that prohibit building user preference models from historical user data, it is getting important to utilize short-term dynamic preferences within a browser session. \textit{Session-based Recommendation} (\textit{SR}) is therefore increasingly used in interactive online computing systems.
      The goal of \textit{SR} is to make recommendations based on user behavior obtained in short web browser sessions, and the task is to predict the users’ next actions, such as clicks, based on previous actions in the same session. 
      
      To better address the \textit{dynamic} nature of \textit{SR} tasks, we study it from an incremental learning perspective, referred to as \textit{Incremental Session-based Recommendation}. In this setting, new items and preferences appear incrementally, and models need to incorporate the new preferences incrementally while preserving old ones that are still useful. The setup requires a recommender to be incrementally updated with data observed during the last time period and evaluated on the events in the following time period. 
      We summarize three main challenges of incremental \textit{SR} scenarios:
   (i). \textit{catastrophic forgetting}~\cite{mccloskey1989catastrophic}:  incorporating new patterns requires additional training and often reduces performance on old patterns.
    (ii). \textit{computation efficiency}: models need to be efficient as they need to be frequently updated with new data.
    (iii). \textit{sample efficiency}: the number of observations on new items and patterns is often small such that models need to capture them quickly with limited observations.

      Recently proposed neural approaches~\cite{hidasi2015session,li2017neural,hidasi2018recurrent,liu2018stamp,sun2019bert4rec} have shown great success in a static setting where the set of items and the distribution of preferences during testing are assumed to be the same as in the training phase. However, this assumption rarely holds in real-world recommendation applications.
As our first contribution, we show that neural recommenders can also be efficiently and effectively used for incremental \textit{SR} scenarios by applying \textit{small} incremental updates. To the best of our knowledge, this is the first study of neural models for \textit{SR} from an incremental learning perspective.
We found that incrementally updating neural models using a single pass with a small learning rate through new data helps to capture new patterns incrementally. However, using large learning rates (or equivalently multiple passes) degrades performance badly due to overfitting new patterns and forgetting old ones.

      Our main contribution is to propose a method called \textbf{M}emory \textbf{A}ugmented \textbf{N}eural model~(\texttt{MAN}) inspired by a framework proposed for language model \cite{merity2016pointer,grave2017unbounded}, neural machine translation~\cite{tu2018learning}, and image recognition \cite{orhan2018simple} tasks.
\texttt{MAN} augments a neural recommender with a \textit{nonparametric} memory to capture new items and preferences incrementally. The predictions of neural and memory components are combined by a lightweight gating network.
\texttt{MAN} is \textit{agnostic} to the neural model as long as it learns meaningful sequence representations, therefore, it can be easily and broadly applied to various neural recommenders.
The nonparametric memory component of \texttt{MAN} helps to deal with all three challenges of incremental \textit{SR} mentioned above.
\textbf{First}, it achieves a long-term memory to remember long histories of observations to mitigate catastrophic forgetting. 
\textbf{Second}, it is very efficient to update because it is not trained jointly with the base neural recommender.
\textbf{Third}, the nonparametric nature by itself helps to better capture new patterns with a smaller number of observations to address the third challenge of sample efficiency. 
Through extensive experiments, we show that \texttt{MAN} boosts the performance of different neural methods and achieves state-of-the-art. We also demonstrate that it is particularly good at capturing preferences on new and infrequent items.

      \section{Related Work}

      
       Incremental learning for recommendation is an important topic for practical recommendation systems where new users, items, and interactions are continuously streaming in. 
    For the standard recommendation task of predicting ratings or clicks, different learners have been studied based on matrix factorization (MF) with
online updating, including \cite{he2016fast}.
      For sequential recommendation tasks, the interaction matrix to be decomposed is constructed from sequential user feedback.
    \cite{rendle2010factorizing} proposes FPMC to factorize the transitions in Markov chains with low-rank representation. 
      Later, \cite{he2016fusing} proposes to FOSSIL with factorized Markov chains to incorporate sequential information.
      Recently, SMF is proposed by~\cite{ludewig2018evaluation} for \textit{SR} using session latent vectors.
      However, these MF-based methods are expensive to train. For example, \cite{li2017neural,wu2018session} reported that even 120GB memory is not enough to train FPMC. Therefore, they are in principle not suitable for incremental settings.

      Session-based recommendation can be formulated as a sequence learning problem to predict the user's sequential behavior. It can be solved by recurrent neural networks~(RNNs).
      The first work (\texttt{GRU4Rec}, \cite{hidasi2015session}) uses a gated recurrent unit (GRU) to learn session representations from previous clicks and predict the next click.
      Based on \texttt{GRU4Rec}, \cite{hidasi2018recurrent} proposes new ranking losses on relevant sessions, and \cite{tan2016improved} proposes to augment training data.
      \texttt{NARM} \cite{li2017neural} augments \texttt{GRU4Rec} with a bilinear decoder and an additional attention operation to pay attention to specific parts of the sequence.  
    Base on \texttt{NARM}, \cite{liu2018stamp} proposes STAMP to model users’ general and short-term interests using two separate attention operations, and \cite{ren2018repeatnet} proposes RepeatNet to use an additional repeat decoder based on an attention mechanism to predict repetitive actions in a session.
    Recently, \cite{wu2018session} uses graph attention to capture complex transitions of items. 
    Motivated by the recent success of \textit{Tansformer} \cite{vaswani2017attention} and \textit{BERT} \cite{devlin2018bert} for language model tasks, \cite{kang2018self} proposed SASRec using the the \textit{Transformer} operation and \cite{sun2019bert4rec} proposed BERT4Rec using the training scheme of \textit{BERT} to model bi-directional information through Cloze tasks.
    Despite the broad exploration and success, these methods are all studied in a static \textit{SR} scenario without considering new items and patterns that appear continuously.
  
    Nonparametric methods~\cite{hardle1994applied} are ideal for incremental learning tasks thanks to their computation efficiency.
      The amount of information they capture increases as the number of observations grows.
    Simple item-based collaborative filtering methods using nearest neighbors have been proven to be effective and are widely employed in industry.
    Markov models \cite{shani2005mdp} also support updating the transition probabilities incrementally. 
      However, the Markov assumption is often too strong, and it limits the recommendation performance. 
      Recently, ~\cite{jannach2017recurrent} proposed \texttt{SKNN} to compare the entire current session with historical sessions in the training data. They show that \texttt{SKNN} is very efficient and achieves strong results. Lately, variations~\cite{ludewig2018evaluation,garg2019sequence} of \texttt{SKNN} have been proposed to consider the position of items in a session or the timestamp of a past session. \cite{mi2017adaptive,mi2018context} built a nonparametric recommender based on a structure called \textit{context tree} to model suffixes of a sequence.
    These simple nonparametric methods are independent from the neural approaches. In this paper, we combine the edges of both neural and nonparametric models for incremental \textit{SR} scenarios.

    Memory-augmented neural models \cite{graves2014neural} have been well-known for the purpose of maintaining long-term memory, which is also recently explored for sequential recommendation tasks \cite{chen2018sequential,ren2019lifelong}.
However, the goal of maintaining long-term memory is not the central concern in incremental \textit{SR} scenarios. This category of memory-augmented neural models differ from ours in the sense that their memory components are \textit{parametric} as they need to be trained jointly with the rest of the model. Therefore, they are not suitable be updated frequently without catastrophic forgetting in incremental \textit{SR} scenarios. 

      Our work is largely inspired by a recently proposed \textit{nonparametric} memory module that is not trained jointly with neural models. \cite{grave2016improving,merity2016pointer} introduce a cache to augment RNNs for language modeling task. They later improve this cache to unbounded size~\cite{grave2017unbounded} and achieve significant performance improvement. 
      A similar memory module is also proposed for neural machine translation tasks~\cite{tu2018learning} and image recognition tasks~\cite{orhan2018simple}.

      \section{Model}

       \paragraph{Background} session-based recommendation, an \textit{event} is a \textit{click} on an item and the task is to predict the next event based on a sequence of events in the current web-browser session.
      Suppose $\mathbf{Y}$ is the set of all candidate items, and the size is $n$.  Existing neural session-based recommenders typically contain two modules: an \textbf{encoder} $g_\theta(\mathbf{x}_{1:t})$ to compute a compact \textit{sequence representation} $\mathbf{c}_t$ of the sequence of events $\mathbf{x}_{1:t}$ until time $t$, and a \textbf{decoder} $      f_{\omega}(\mathbf{c}_t)$ to compute an output distribution $P^N \in \mathbb{R}^n$ to predict the next event.
      Recurrent neural networks (RNNs) and fully-connected layers are common choices for encoder and decoder respectively. 
    In our later experiments, we included models that use different types of encoders and decoders.
    Our \texttt{MAN} framework is agnostic to the neural recommender, therefore, readers can use many other neural architectures with an encoder-decoder structure. 
      
      Next, we present the \textbf{M}emory \textbf{A}ugmented \textbf{N}eural recommender (\texttt{MAN}) to augment a neural session-based recommender with a cache-like nonparametric memory to incrementally incorporate new items and preferences.
      We first introduce the architecture of the cache-like memory. Then we describe how to use it to generate nonparametric memory predictions, and how to merge memory and neural predictions through a lightweight gating network.
      
      \subsection{Nonparametric Memory Structure}
      Our memory $\mathbf{M} $ is an
      array of slots in the form of (key, value) pairs \cite{miller2016key} in Eq.(\ref{eq:memory}). $\mathbf{M}$ is queried by keys and returns corresponding values.
      We define the keys to be input sequence representations $\mathbf{c}_m$ computed by the neural encoder, and the values to be the corresponding label $y_m$ of the next event.
    To scale to a large set of observations and long histories in practical recommendation scenarios, we do not restrict the size of $\mathbf{M}$ but store all pairs $(\mathbf{c}_m, y_m)$ from previous events. 
    
      \begin{equation}
      \mathbf{M} = \{(\mathbf{c}_1, y_1), \dots, (\mathbf{c}_m, y_m), \dots\}
    \label{eq:memory}
      \end{equation}    
    
      \subsection{Memory Prediction}
      
      To support efficient incremental learning, the memory module is \textit{not} trained jointly with the neural recommender. Instead, it directly predicts the next event by computing a probability distribution using entries stored in $\mathbf{M}$.
      For an input sequence $\mathbf{x}_{1:t}$, we first match the current sequence representation $\mathbf{c}_t = g_\theta(\mathbf{x}_{1:t})$ against $\mathbf{M}$ to retrieve $K$ nearest neighbors $\mathcal{N}({\mathbf{c}_t}) = \{(\mathbf{c}_k, y_k, d_k)\}_{k=1}^{K}$ of $\mathbf{c}_t$,   
      where $d_k$ is the Euclidean distance between the sequence representation $\mathbf{c}_k$ of the $k$-th neighbor and $\mathbf{c}_t$.
      Then, we use the $K$ nearest neighbors to compute a non-parametric memory prediction $P^{M} \in \mathbb{R}^n $ using a variable kernel density estimation by: 
      \begin{equation}
      P^{M}(y_i) \propto \sum_{k=1}^{K} \delta(y_i = y_k) \mathbf{K} \left( \dfrac{ d_k }{ d^\star } \right)
      \label{eq:mempred}
      \end{equation}    
      where $P^{M}(y_i)$ is the probability on an item $y_i \in \mathbf{Y}$, $\delta(y_t = y_k)$ is Kronecker delta which equals one when the equality holds and zero otherwise, $\mathbf{K}$ is a Gaussian kernel ($\mathbf{K}(x) = exp(-x^2/2)$), and $d^\star$ is the Euclidean distance between $\mathbf{c}_{t}$ and its closest neighbor in $\mathcal{N}({\mathbf{c}_t})$.
    $P^M$ only assigns non-zero probability to at most $K$ (number of neighbors) items because the probabilities assigned to items that do not appear in the nearest neighbors are zero. As a result, $P^{M}$ is a very \textit{sparse} distribution as a mixture of the labels in  $   \mathcal{N}({\mathbf{c}_t})$ weighted by their similarities to $\mathbf{c}_t$. 
     To capture new preference patterns incrementally, $\mathbf{M}$ is queried and updated incrementally during the testing phase such that $P^{M}$ is up-to-date.
      
               \begin{figure*}[htb!]
            \centering
            \includegraphics[width=0.92\textwidth]{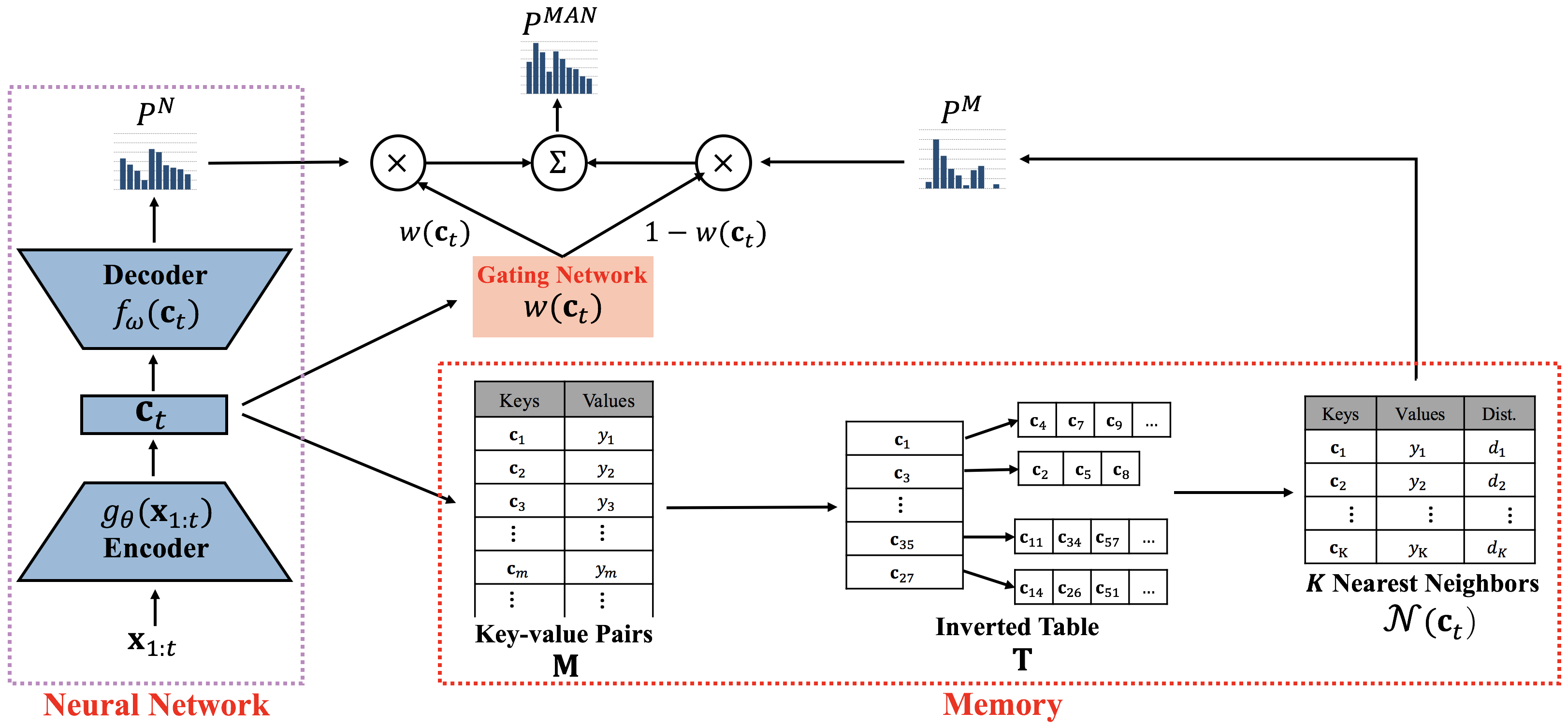}
            \vspace{-0.05in}\caption{Computation pipeline of \textbf{M}emory \textbf{A}ugmented \textbf{N}eural recommender (\texttt{MAN}). Predictions from a neural network are augmented by predictions from a memory module through a gating network. }
            \label{fig:model}
      \end{figure*}

      \subsection{Gating Network to Combine Memory and Neural Predictions}
    
      The neural prediction $P^{N}$ mainly captures static old preference patterns while the nonparametric memory prediction $P^{M}$ can model infrequent and new preference patterns incrementally. To flexibly work with both scenarios, these two predictions are combined.
A simple way proposed by \cite{grave2017unbounded,orhan2018simple} is linearly interpolating them with a fixed weight, and we later call this version ``\texttt{MAN-Shallow}'' in our experiments in Section \ref{subsec:analysis}. 

To better merge these two predictions at different sequential contexts, we propose to use a lightweight \textbf{gating network} \cite{bakhtin2018lightweight} $w(\mathbf{c}_t)$ to learn the mixing coefficient as a function of the sequence representation $\mathbf{c}_t$.
We use a lightweight fully connected neural network defined in Eq.(\ref{eq:gating}) with a single hidden layer of 100 hidden units, \textit{tanh} as hidden layer activation function, and a Sigmoid function at the output layer. 
\begin{equation}
w(\mathbf{c}_t) = \sigma(\mathbf{W}_o * tanh(\mathbf{W}_h \mathbf{c}_t + \mathbf{b}_h) +b_o)
\label{eq:gating}
\end{equation}
The output of $w(\mathbf{c}_t)$ is a scalar between 0 and 1 that measures the relative importance of $P^{N}$, while the other $1-w(\mathbf{c}_t)$ fraction is multiplied to $P^{M}$.
The final prediction distribution $P^{MAN}$ is an learning interpolation of $P^M$ and $P^N$ weighted by the output of $w(\mathbf{c}_t)$ computed in Eq.(\ref{eq:final}) .
      \begin{equation}
      P^{MAN}(y_t)  =  w(\mathbf{c}_t) P^{N}(y_t) +   (1 - w(\mathbf{c}_t) ) P^{M}(y_t)
      \label{eq:final}
      \end{equation}
The gating network is trained with cross-entropy loss using $P^{MAN}$ as predictive distribution \textit{after} the normal training of the neural model with both $P^{N}$ and $P^{M}$ are fixed. The gradients are not computed for the large number parameters of the neural model to avoid computation overhead and interference with the trained neural model.
Inspired by \cite{bakhtin2018lightweight}, it is trained using only validation data. The idea is to train it using data not seen during the pre-training phase to better predict new preferences that might appear in incremental \textit{SR} scenarios. We randomly select 90\% validation data for training and the remaining 10\% for early stopping.
We compared this setup with training using the whole training set, and we found that our setup achieves better performance while being much more efficient.


      \subsection{Efficient Large-scale Nearest Neighbor Computation}
      
      As the number of events in practical \textit{SR} scenarios is huge and we do not restrict the size of $\mathbf{M}$, computing nearest neighbors frequently to generate $P^{M}$ can be expensive.
   We apply a scalable retrieval method used by \cite{grave2017unbounded}.
      To avoid exhaustive search, an \textit{inverted file table} $\mathbf{T}$ is maintained. Keys in $\mathbf{M}$ are first clustered to a set of clusters using k-means, then all keys in  $\mathbf{M}$ can be associated with one centroid. When we query $\mathbf{c}_t$ in $\mathbf{M}$, it is searched by firstly matching to a set of centroids to get the closest cluster and then the set of keys in this cluster.
      
The clustered memory supports efficient querying, yet it is memory consuming because each key in $\mathbf{M}$ needs to be stored. This can be greatly reduced by Product Quantization (PQ~\cite{jegou2010product}) that quantizes a vector by parts (sub-quantizers), and it does not directly store the vector but its residual, i.e., the difference between the vector and its associated centroids.
    We use $2^8$ centroids, 8 sub-quantizers per sequence representation, and 8 bits allocated per sub-quantizer, then we only need the size of 16 (quantization code + centroid id = 8 + 8) bytes per vector.  Therefore, a million sequence representations can be stored with only 16 Mb memory.
      With an inverted table and PQ, we have a fast approximate nearest neighbor retrieval method with a low memory footprint. We use the FAISS \footnote{\url{https://github.com/facebookresearch/faiss}} open-source library that also supports GPU acceleration for implementation.
         
        \begin{algorithm}[t]
            \begin{algorithmic}[1]
                  \Procedure{Train}{$ D_{train}, D_{valid}, \mathbf{M}$}
                  \State Train $g_\theta$, $f_{\omega}$ w.r.t. $D_{train}$
                  \For {$ (\mathbf{x}_{1:t}, y_t) \in D_{train}$}
                  \State Compute $\mathbf{c}_t = g_\theta(\mathbf{x}_{1:t})$; store $(\mathbf{c}_t, y_t)$ to $\mathbf{M}$ 
                  \EndFor
                  \State Build the inverted table $\mathbf{T}$ for $\mathbf{M}$ 
            \State Fix $g_\theta$, $f_{\omega}$, and train $w(\mathbf{c}_t)$ w.r.t. $D_{valid}$
                  \EndProcedure
                  \Procedure{Test}{$ D_{test}, \mathbf{M}$}
                  \For {$ (\mathbf{x}_{1:t}, y_t) \in D_{test}$}
                  \State Compute $\mathbf{c}_t = g_\theta(\mathbf{x}_{1:t})$, and $P^{N} = f_{\omega}(\mathbf{c}_t)$
                  \State Query $\mathbf{M}$ and $\mathbf{T}$ to retrieve $ \mathcal{N}({\mathbf{c}_t})$
                  \State Compute $P^M$ by Eq.(\ref{eq:mempred}), $P^{MAN} $ by Eq.(\ref{eq:final})
            \State Update $\mathbf{M}$ and $\mathbf{T}$ with $(\mathbf{c}_t, y_t)$
                  \EndFor
                  \EndProcedure
            \caption{\textbf{M}emory \textbf{A}ugmented \textbf{N}eural Recommender}
                  \label{alg:one}
            \end{algorithmic} 
      \end{algorithm}

\subsection{Overall \texttt{MAN} Algorithm}

\label{subsec:algo}

The computation pipeline and algorithm of \texttt{MAN} presented in Figure~\ref{fig:model} and Algorithm 1. Next, we describe the training and testing procedures of \texttt{MAN} in detail, and also analyzed its computation efficiency.
   
      \paragraph{Training procedure} \texttt{MAN} first trains the neural encoder and decoder on the training set $D_{train}$. Then, it computes sequence representations for all training data and stores them with corresponding labels to $\mathbf{M}$. Afterwards, the clustered memory of the inverted table $\mathbf{T}$ is built with entries in $\mathbf{M}$.
      Lastly, The gating network is trained on validation set $D_{valid}$.
      
     \paragraph{Testing procedure} The sequence representation $\mathbf{c}_t$ and the neural prediction $P^{N}$ is first computed. Then, $\mathbf{M}$ and $\mathbf{T}$ are queried with $\mathbf{c}_t$ to retrieve $K$ nearest neighbors $ \mathcal{N}({\mathbf{c}_t})$ to compute $P^{M}$ . 
      To generate the final recommendation $P^{MAN}$, $P^{M}$ is merged with $P^{N}$ weighted by the output of the gating network. 
    Lastly, $\mathbf{M}$ and $\mathbf{T}$ are incrementally updated with the new testing pair $(\mathbf{c}_t, y_t)$.
During testing, the clustered memory in $\mathbf{T}$ is not updated for the purpose of computation efficiency. Running k-means to update the clustered memory on huge data sets with large dimensions is computationally intensive. Therefore, we need to decide when and how to update the clustering algorithm, and there
will be a trade-off between the performance benefits and the
computation overhead. Studies on this part are left for interesting future work.

            \begin{table*}
            \centering
            \tabcolsep=0.25cm
            \begin{tabu}{c rrr rr rrc}
                  \toprule
                  & \multicolumn{3}{c}{Training Data} & \multicolumn{2}{c}{Validation Data} & \multicolumn{3}{c}{Testing Data} \\
                  \cmidrule(r){2-4} \cmidrule(r){5-6} \cmidrule(r){7-9}
                  & Events &  Sessions & Items &   Events &  Sessions &  Events &  Sessions & New Events \\
                  \textbf{YOOCHOOSE} & 6,245,412 & 1,535,693 & 22,594 & 693,935 & 170,633  & 748,269 & 178,920  & 8.6\%\\
                  \textbf{DIGINETICA} & 636,506 & 130,994 & 42,294 & 70,723 & 14,555  & 286,254 & 59,240  & 3.3\% \\
                  \bottomrule             
            \end{tabu}
          \caption{ Statistics of two datasets. The last column (`New Events') indicates the percentage of testing events that involve new items not in the training set.}  \vspace{-0.08in}
            \label{table:statistics}
      \end{table*}
    
    \paragraph{Computation efficiency analysis}
    During training, the additional training procedures of \texttt{MAN} on top of the regular neural recommender training are efficient. (i) Sequence representations of training data can be obtained directly from the last regular training epoch of the neural recommender, therefore, no computation overhead is injected at this step (line 3-5). 2). (ii) Building the clustered memory and the inverted table for entries in $\mathbf{M}$ (line 6) is also fast with the FAISS library. (iii) Training the lightweight gating network using only validation split (line 7) is much more efficient than the regular training of the base neural recommender.
During testing, querying the memory to retrieve nearest neighbors can be done very efficiently supported by FAISS; a forward computation through the lightweight gating network is also efficient.



      \section{Experiments and Analyses}

      \subsection{Datasets}
      
      \quad \textbf{YOOCHOOSE:} This is a public dataset for RecSys Challenge 2015.\footnote{\url{http://2015.recsyschallenge.com/challenge.html}} It contains click-streams on an e-commerce site over 6 months. Events in the last week are tested. 
      Following~\cite{tan2016improved,li2017neural}, we use the latest 1/4 training sequences because YOOCHOOSE is very large.
      
      \textbf{DIGINETICA:}
      This dataset contains click-streams data on another
      e-commerce site over a span of 5 months for CIKM Cup 2016.\footnote{\url{http://cikm2016.cs.iupui.edu/cikm-cup}}
      Events in the last four weeks are tested.
      
       Items that appear less than five times, and sessions of length shorter than two or longer than 20 are filtered out.
      Statistics of the two datasets after pruning are summarized in Table~\ref{table:statistics}, and the last 10\% of the training
data based on time is used as the validation set.
      Different from previous static settings that remove items not in the training phase from test sets, our test sets for the incremental \textit{SR} task include events on new items that appear only during the testing phase. 
    The last column indicates the percentage of events in the test data that involve items not part of the training data.

      \subsection{Evaluation Metrics}

        \quad \textbf{HR@k}: Average hit rate when the desired item is
            amongst the top-k recommended items. It can be interpreted as precision~\cite{liu2018stamp,wu2018session} or recall~\cite{hidasi2015session,li2017neural,jannach2017recurrent} because we predict the immediate next event.
        
            \textbf{MRR@k}: HR@k does not consider the order of the items recommended. MRR@k measures the mean reciprocal ranks of the desired items in top-k recommended items.
    
      \subsection{Models and Training Details}
      
      \textbf{\quad \texttt{Item-KNN}}: 
      This simple baseline recommends top items similar to the single last item in the current session based on co-occurrence statistics of two items in other sessions.
      
      \textbf{\texttt{(S)-SKNN}}:
      Instead of considering only the last item, \texttt{SKNN}~\cite{jannach2017recurrent} compares all items in the current session with items in other sessions. \texttt{S-SKNN} \cite{ludewig2018evaluation} is an improved version that assigns more weights to items that appear later in a session.
      
       \textbf{\texttt{CT}}~\cite{mi2018context}: It builds a nonparametric recommender based on a structure called Context Tree to model suffixes of a sequence. 
      
      \textbf{\texttt{GRU4Rec}}~\cite{hidasi2018recurrent}: It uses an RNN with GRU as encoder. It also uses specialized ranking-based losses computed w.r.t. the most relevant samples, which perform better than their initial version \cite{hidasi2015session}.
      
      \textbf{\texttt{NARM}}~\cite{li2017neural}: 
      It improves the encoder of \texttt{GRU4Rec} with an item-level attention and replaces the decoder by a bilinear decoding scheme. 
    
      
      \textbf{\texttt{MAN} (proposed):} The method proposed in this paper. Two versions  (\texttt{MAN-GRU4Rec} and \texttt{MAN-NARM}) are tested using \texttt{GRU4Rec} and \texttt{NARM} as base neural models respectively. Unless mentioned specifically, \texttt{MAN} is based on \texttt{NARM}. \footnote{In this  paper, we chose the two most representative base neural models (\texttt{GRU4Rec} and \texttt{NARM}) to evaluate \texttt{MAN}. As \texttt{MAN} is agnostic to the neural model, the exploration of using \texttt{MAN} on top of other neural recommenders is left for future work.}

    \begin{table}
     \tabcolsep=0.064cm
            \centering
            \begin{tabular}{ccccccccc}
                  \toprule
                  \textbf{$\eta$} & 1e-2 & 5e-3 & 1e-3 & 5e-4 & 1e-4 &5e-5&0\\
                  \midrule
            \small \textbf{YOOCHOOSE} & & & & & &\\
                  \texttt{NARM}   &0.389 & 0.422 & 0.460 & \textbf{0.463} & 0.447 & 0.440 & 0.420\\
           \small \textbf{DIGINETICA} & & & & & &\\
                  \texttt{NARM}  & 0.235 & 0.255 & 0.315  & 0.338& \textbf{0.355} & 0.350 & 0.324\\
                  \bottomrule
            \end{tabular}
            \caption{HR@5 with different learning rate $\eta$ to update \texttt{NARM} incrementally. \texttt{NARM} is fixed during testing when $\eta=0$.}  \vspace{-0.05in}
            \label{table:learningrate}
      \end{table}

            \begin{table*} 
            \centering
        \tabcolsep=0.28cm
            \begin{tabular}{l  cccc  cccc }
                  \toprule
                  & \multicolumn{4}{c}{\textbf{YOOCHOOSE}}      & \multicolumn{4}{c}{\textbf{DIGINETICA}} \\
                  \cmidrule(r){2-5} \cmidrule(r){6-9}
                  & HR@5  & MRR@5 & HR@20 & MRR@20 & HR@5  & MRR@5 & HR@20 & MRR@20  \\
                  \texttt{Item-KNN}                    &     0.205  &    0.114   &   0.403    &   0.127     &   0.112    &   0.042    &   0.186    &   0.056    \\
                  \texttt{GRU4Rec}               & 0.359 & 0.216 & 0.582 & 0.228  & 0.191 & 0.114 & 0.382 & 0.135   \\
                  \texttt{SKNN}                  & 0.411 & 0.245 & 0.625 & 0.268  & 0.262 & 0.156 & 0.489 & 0.177   \\
            \texttt{S-SKNN}                  & 0.416 & 0.247 & 0.628 & 0.272  & 0.279 & 0.170 & 0.497 & 0.185   \\
            \texttt{CT}                  & 0.427 & 0.263 & 0.618 & 0.286  & 0.290 & 0.189 & 0.515 & 0.206   \\
                  \texttt{MAN-GRU4Rec}           & 0.447 & 0.269 & 0.657 & 0.293  & 0.331 & 0.203 & 0.545 & 0.226  \\
            \texttt{NARM}                  & 0.463 & 0.280 & 0.682 & 0.303  & 0.358 & 0.221 & 0.566 & 0.242  \\
            \texttt{MAN-NARM}           & \textbf{0.476} & \textbf{0.292} & \textbf{0.689} & \textbf{0.314}  & \textbf{0.381} & \textbf{0.234} & \textbf{0.599} & \textbf{0.258}  \\
                  \bottomrule
            \end{tabular}
        \caption{Overall results of different models for the incremental \textit{SR} task on two datasets. Models are ranked by HR@5, and the best method in each column is in bold.}
       
            \label{table:results}
      \end{table*}
       
        \begin{figure*}[htb]
            \centering
            \includegraphics[width=0.48\textwidth]{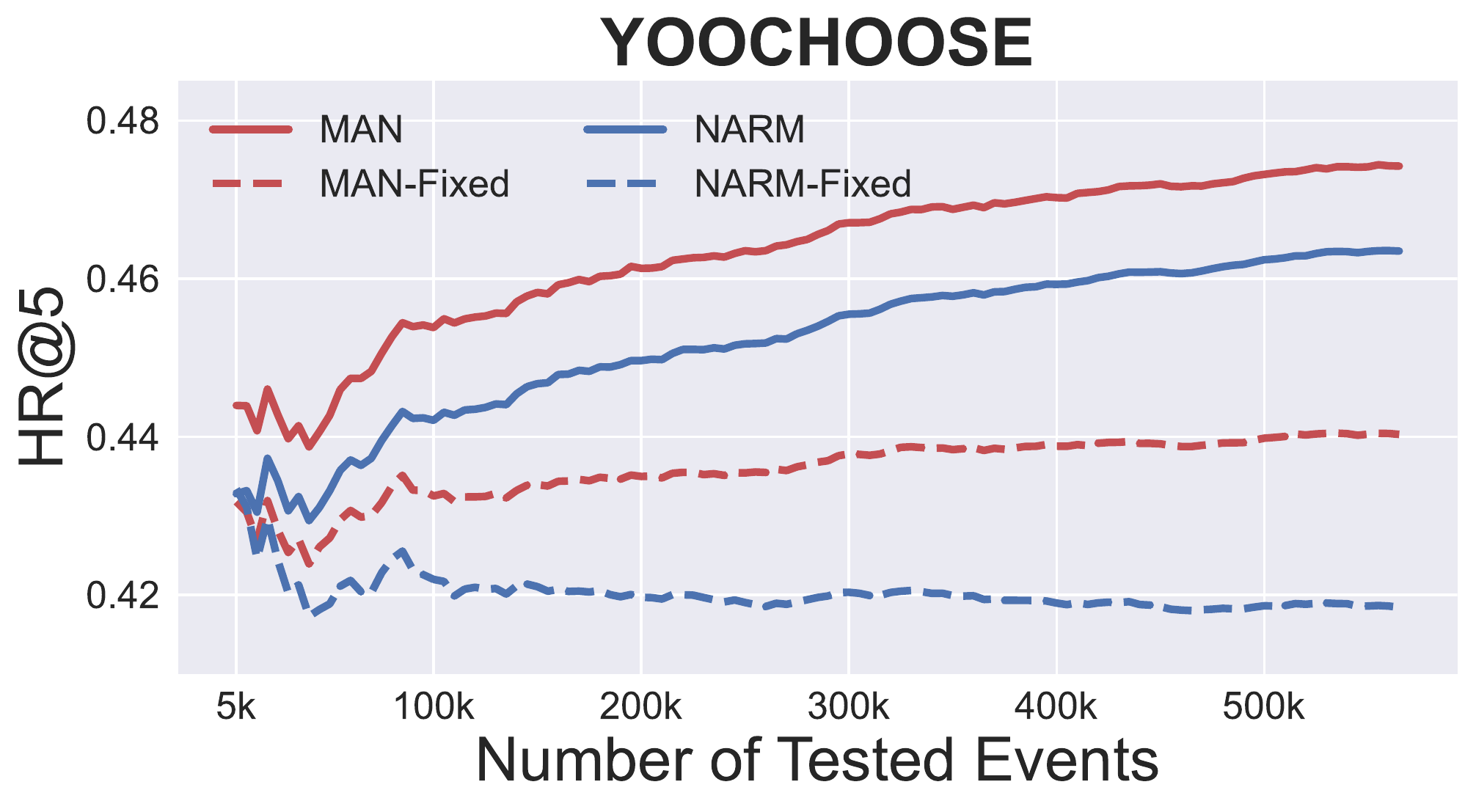}  \quad
            \includegraphics[width=0.48\textwidth]{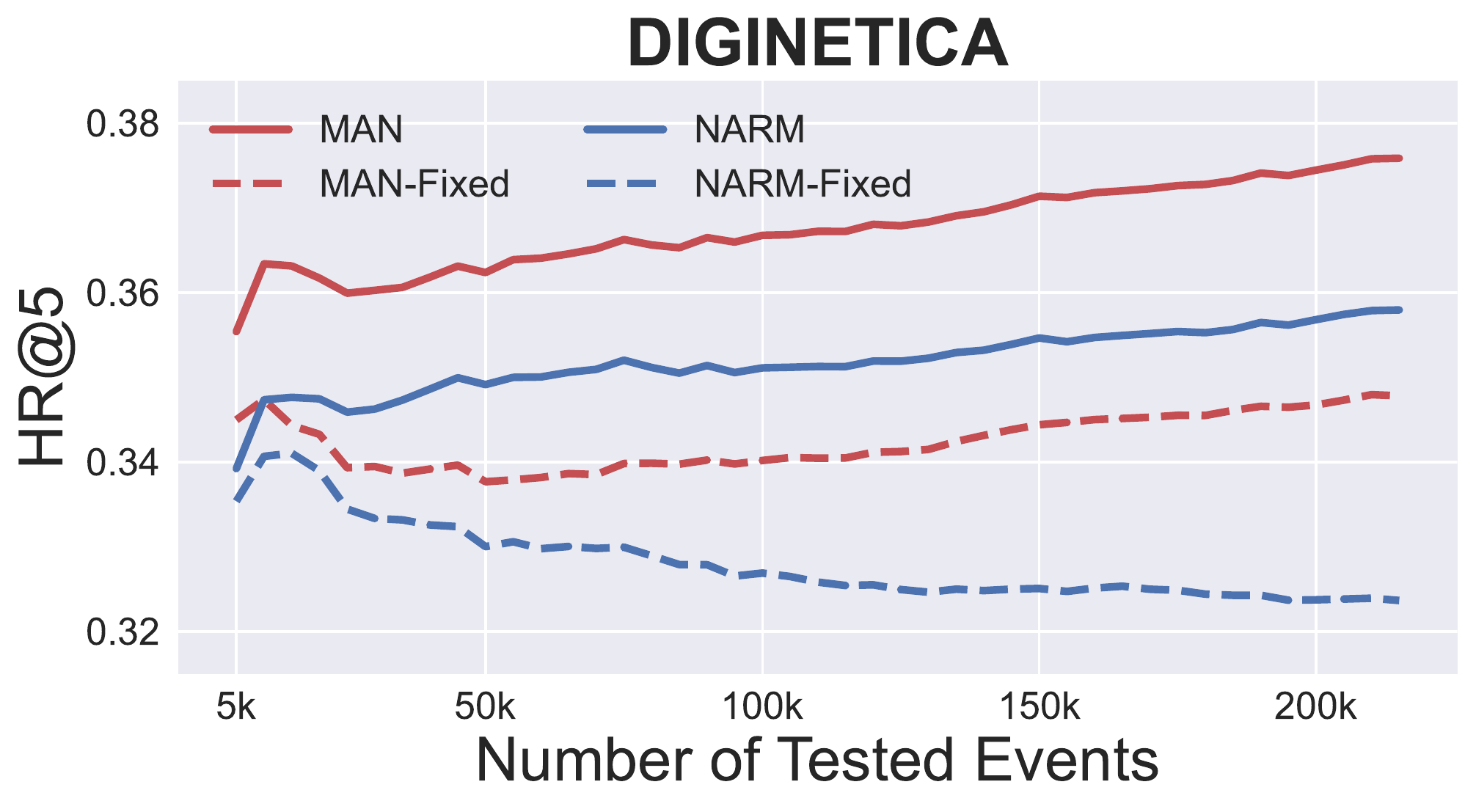}
           \vspace{-0.01in}\caption{Incremental performance (HR@5) as the number of tested events increases.} \vspace{-0.01in}
            \label{fig:curve}
      \end{figure*} 
During  training, hidden layer size of \texttt{GRU4Rec} and \texttt{NARM} is set to 100, and the item embedding size of \texttt{NARM} is set to 50; the batch size is set to 512 and 30 epochs are trained for both models. 
Hyper-parameters of different models are tuned on validation splits to maximize HR@5.
The number of nearest neighbors of \texttt{MAN} is set to 50 for YOOCHOOSE and 100 for DIGINETICA.
During testing, we update different neural models and \texttt{MAN} incrementally using \textit{a single gradient descent step} as every batch of 100 events are tested. Learning rates for neural models are 5e-4 and 1e-4 for YOOCHOOSE and DIGINETICA, and the learning rate to update the gating network is 1e-3. 
Other pure nonparametric methods (\texttt{Item-KNN}, \texttt{(S)-SKNN}, \texttt{CT}) are also incrementally updated as every batch of 100 events is tested.

      \subsection{Experiment Results}
  
  In this section, we first study the effect of using different learning rates to update neural models incrementally. Then, we analyze the overall performance of different methods and plot the incremental performance of two representative methods as more events are tested.
  
  \paragraph{Effect of incremental learning rate} Before we proceed to our main results, we first highlight that the degree of updates is important when neural models are updated incrementally.
As an example, we show results of using different learning rates to update \texttt{NARM} in Table \ref{table:learningrate}. 
   We can see that using large learning rates \footnote{
   Using multiple incremental training epochs has a similar effect as using large earning rates in one epoch.} to incrementally update \texttt{NARM} \textit{degrades} performance severely, while using relatively small learning rates outperforms the version when \texttt{NARM} is fixed during testing ($\eta=0$). 
   We believe that large learning rates cause the model to overfit new patterns while catastrophically forgetting old patterns. 
   Therefore, we contend that \textit{incremental updates for neural models needs to be small}.
   
      \paragraph{Overall Performance}
      The results of different methods on two datasets are summarized in Table~\ref{table:results}. 
      Several interesting empirical results can be noted:
      \textbf{First}, \texttt{MAN-NARM} achieves the top performance.
Both \texttt{MAN-NARM} and \texttt{MAN-GRU4Rec} consistently outperform their individual neural modules~(\texttt{NARM} and \texttt{GRU4Rec}) with notable margins. This result shows that our \texttt{MAN} architecture effectively helps standard neural recommenders for incremental \textit{SR} scenarios.
      \textbf{Second}, \texttt{MAN} is not sensitive to neural recommenders. We observed that even though \texttt{NARM} significantly outperforms \texttt{GRU4Rec}, \texttt{MAN-GRU4Rec} and \texttt{MAN-NARM} show comparable performances. We contend that the memory predictions effectively compensate the failed predictions of \texttt{GRU4Rec}.
    \textbf{Third}, \texttt{(S)-SKNN} and \textsc{CT} are much stronger nonparametric methods than \texttt{Item-KNN}, with \textsc{CT} being slightly better. They both outperform \texttt{GRU4Rec} on two datasets, and their performance gaps compared to \texttt{MAN} and \texttt{NARM} are smaller on the YOUCHOOSE dataset that contains more new events.

 \paragraph{Incremental Performance}

      In Figure~\ref{fig:curve}, we present incremental performance of \texttt{MAN} and \texttt{NARM} as more events are evaluated during testing. We also included versions that are \textit{fixed} during the testing phase.
    In \texttt{NARM-Fixed}, the neural model is fixed. In \texttt{MAN-Fixed}, both the neural model and the gating networks are fixed, and only memory entries are incrementally expanded.
    Several results can be noted from Figure~\ref{fig:curve}.
    \textbf{First}, \texttt{MAN} helps to boost performance both when the neural model is fixed and incrementally updated. \texttt{MAN} outperforms \texttt{NARM} by consistent margins, and \texttt{MAN-Fix} outperforms \texttt{NARM-Fix} by increasing margins.
    \textbf{Second}, incrementally updated models consistently outperform their fixed versions. Both \texttt{MAN} and \texttt{NARM} outperform their corresponding fixed versions with significant and increasing margins.

  \subsection{In-depth Analysis} 
  \label{subsec:analysis}
 In the following experiments, we conduct in-depth analysis to further understand the effectiveness and efficiency of different components of \texttt{MAN} and for what types of events \texttt{MAN} most improves predictions.

      \begin{table}
      \centering
      \tabcolsep=0.15cm
            \begin{tabular}{lcccc}
                  \toprule
                  & \multicolumn{2}{c}{\textbf{YOOCHOOSE}} & \multicolumn{2}{c}{\textbf{DIGINETICA}} \\
                  \cmidrule(r){2-3} \cmidrule(r){4-5}
                  & HR@5         & MRR@5        & HR@5         & MRR@5        \\
                  \texttt{MAN}        & \textbf{0.476 }        & \textbf{0.292}         & \textbf{0.381}         & \textbf{0.234}  \\
                  \texttt{MAN-Shallow} & 0.469         & 0.286         & 0.374        & 0.228          \\
                  \texttt{MAN-50k} & 0.469         & 0.287         & 0.376         & 0.231          \\
                  \texttt{MAN-10k} & 0.466         & 0.284         & 0.369         & 0.226          \\
                  \bottomrule       
            \end{tabular}     
        \caption{Ablation study for \texttt{MAN}. \textbf{The large memory size and the gating network of \texttt{MAN} gain performance benefits.}} 
            \label{table:ablation}
      \end{table}

      \paragraph{Ablation Study}
  
   We further studied in Table~\ref{table:ablation} the effect of two setups in \texttt{MAN}, i.e., the large memory size and the combining scheme with a gating network. Two simpler versions are compared. \textbf{(i)} \texttt{MAN-Shallow} linearly combines neural and memory predictions with a fixed scalar rather than the weight output by the gating network. The scalar weight is tuned on validation splits to be 0.7 for YOOCHOOSE and 0.8 for DIGINETICA. This simple method serves as a strong baseline in learning new vocabularies in language model tasks \cite{merity2016pointer,grave2017unbounded}.
\textbf{(ii)} \texttt{MAN-50k/10k} use fixed-size memories that only store a limited number of recent events (50k/10k).
      \texttt{MAN-Shallow} is consistently inferior to \texttt{MAN}. It means the gating network makes a better decision to combine neural and memory predictions.
    Furthermore, the performance drops as the size of the memory decrease from unbounded (\texttt{MAN}) to 50k, and to 10k. It means that keeping a big memory that handles long histories achieves better recommendation performance. Despite the slight performance drop, the three simplified versions are more efficient, therefore, they are still suitable candidates for industry.

\begin{figure}[t]
		\centering
		\includegraphics[width=0.48\textwidth]{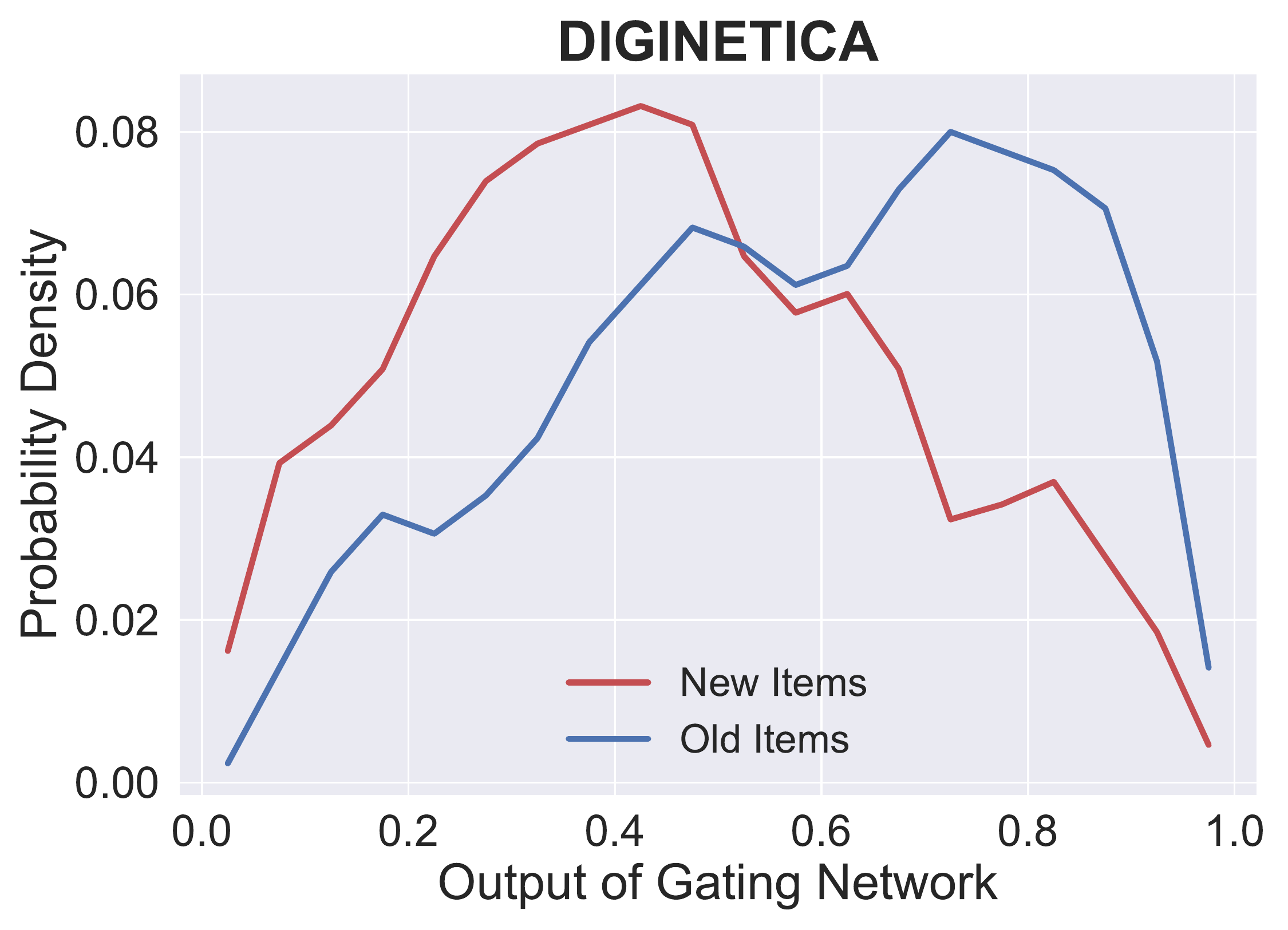} 
    \vspace{-0.15in} \caption{  Visualizing the empirical distribution of the Gating Network outputs assigned to the parametric part on DIGINETICA. Outputs on old (in the training phase) and new (not in the training phase) items are plotted separately. \textbf{\texttt{MAN} assigns small weights to the parametric part when items are new, and large weights to it when items are old}. } \vspace{0.1in}
        \label{fig:output}
\end{figure}

\begin{table}[t]
\centering
\setlength{\tabcolsep}{4pt}
\begin{tabular}{ccccc}
\toprule
& \multicolumn{2}{c}{\textbf{YOOCHOOSE}} & \multicolumn{2}{c}{\textbf{DIGINETICA}} \\
\cmidrule(lr){2-3}  \cmidrule(lr){4-5} 
& Train  & Avg. Response & Train & Avg. Response \\
\cmidrule(lr){2-3}  \cmidrule(lr){4-5} 
NARM & 375.5 & 25.5 & 42.5 & 23.5 \\ 
MAN & 392.2 & 29.5 & 45.8 & 27.8 \\ 
\bottomrule
\end{tabular}
\caption{Computation time of \texttt{MAN} and \texttt{NARM}. Total training time (in minutes) and average response time (in milliseconds) per recommendation are reported for both datasets. \textbf{The computation overhead of \texttt{MAN} on top of \texttt{NARM} is trivial}.}
\label{table:time}
\end{table}

    \paragraph{Visualization of Gating Network Outputs}
    
In this experiment, we further visualize the empirical distribution of the gating network outputs assigned to the parametric part. We demonstrate in Figure \ref{fig:output} that it indeed learns desired weight distributions to the parametric and non-parametric parts.
For new items denoted by the red curve, outputs of the gating network tend to be small such that larger weights are assigned to the non-parametric memory prediction.
Similarly, for old items denoted by the blue curve, outputs of the gating network tend to be large such that smaller weights are given to the non-parametric memory prediction.

\paragraph{Computation Time}

In Table \ref{table:time}, we report the total training time and average response time per recommendation of \texttt{MAN} compared to \texttt{NARM}. Both models are trained using a NVIDIA TITAN X GPU with 12GB memory. We can see that the extra training time and response time of \texttt{MAN} is \textbf{trivial} on top of the \texttt{NARM}. 
This empirical results, together with our analysis in Section \ref{subsec:algo}, demonstrate that \texttt{MAN} is computation efficient.


      \paragraph{Disentangled Performance}

      To further understand for what types of events that \texttt{MAN} most improves predictions, we studied the disentangled performance when items are bucketed into five groups by their occurrence frequency in training data. Bucket 1 contains the least frequent items, and bucket 5 contains the most frequent items. Bucket splitting intervals are chosen to ensure the bucket size is the same. 
     The disentangled performances of \texttt{MAN} and \texttt{NARM} across five buckets are reported in Table \ref{table:bucket}, and the results reveal that:
     \begin{itemize} 
     \item  Infrequent items are more challenging to predict. The performances of both methods have an increasing trend as item frequency increases.
\item  \texttt{MAN} is consistently better than \texttt{NARM} on all levels of item frequency.
\item  The improvement margin of \texttt{MAN} over \texttt{NARM} is very significant on infrequent items (small bucket number). Therefore, we contend that \texttt{MAN} it is especially good at learning new patterns and items with a small number of observations.
     \end{itemize}
            \setlength{\tabcolsep}{5pt}
    \begin{table} [t]
            \centering
            \begin{tabular}{ccccccc}
                  \toprule
                  \textbf{Bucket \#} & 1  & 2 & 3 & 4 & 5\\
                  \midrule
            \small \textbf{YOOCHOOSE} & & & & \\
                  \texttt{MAN} & 0.318  &0.263 & 0.319 & 0.378 & 0.517 \\
                  \texttt{NARM} & 0.287  &0.247 & 0.302 & 0.362 & 0.510\\
           \small \textbf{DIGINETICA} & & & & \\
                  \texttt{MAN} & 0.145  & 0.187 & 0.251 & 0.329  & 0.515\\
                  \texttt{NARM} & 0.117  & 0.169 & 0.230 & 0.303  & 0.494\\
                  \bottomrule
            \end{tabular}
            \caption{Disentangled performance (HR@5) at different item frequency buckets. \textbf{\texttt{MAN} is consistently better at all buckets, and the improvement margin on infrequent items is the most significant.}} 
            \label{table:bucket}
      \end{table}

      \section{Conclusion}

      In this paper, we study neural methods in a realistic incremental session-based recommendation scenario. 
      We show that existing neural models can be used in this scenario with small incremental updates, and we propose a general method called \textbf{M}emory \textbf{A}ugmented \textbf{N}eural recommender (\texttt{MAN}) that is widely applicable to augment different existing neural models.
      \texttt{MAN} uses a efficient nonparametric memory to compute a memory prediction, which is combined with the neural prediction through a lightweight gating network.
      We show that \texttt{MAN} consistently outperforms state-of-the-art neural and nonparametric methods and it is particularly good at learning new items with insufficient number of observations.
   
     \small
      \bibliographystyle{named}
      \bibliography{rec}
      
\end{document}